\documentclass[10pt,twocolumn,letterpaper]{article}

\usepackage{cvpr}
\usepackage{times}
\usepackage{epsfig}
\usepackage{graphicx}
\usepackage{amsmath}
\usepackage{amssymb}
\usepackage{multirow}
\usepackage{booktabs}
\usepackage{algorithm}
\usepackage{algorithmic}
\usepackage{arydshln}
\usepackage{threeparttable}
\usepackage{colortbl}
\usepackage{enumitem}
\usepackage{bm}
\usepackage[numbers]{natbib}
\usepackage{array}
\usepackage[table]{xcolor}
\usepackage{colortbl}
\usepackage{caption}
\usepackage{dblfloatfix}
\definecolor{mygray}{gray}{.9}
\definecolor{ggray}{RGB}{127,127,127}
\definecolor{reda}{RGB}{192,0,0}
\definecolor{redb}{RGB}{217,148,143}
\definecolor{myyellow}{RGB}{190,144,0}
\definecolor{mygreen}{RGB}{80,100,40}
\definecolor{myblue}{RGB}{30,90,100}
\newcommand{\tabincell}[2]{\begin{tabular}{@{}#1@{}}#2\end{tabular}}
\makeatletter
\newcommand{\thickhline}{%
	\noalign {\ifnum 0=`}\fi \hrule height 1pt
	\futurelet \reserved@a \@xhline
}
\usepackage[pagebackref=true,breaklinks=true,letterpaper=true,colorlinks,bookmarks=false]{hyperref}
\usepackage{pifont}
\usepackage[breaklinks=true,bookmarks=false]{hyperref}

\cvprfinalcopy 

\def\cvprPaperID{1308} 

\ifcvprfinal\fi
\begin{document}

\title{Learning Video Object Segmentation from Unlabeled Videos}

\author{Xiankai Lu$^{1}$~\!,\hspace{1pt} Wenguan Wang$^{2}\thanks{Corresponding author: \textit{Wenguan Wang}.}$~\!, \hspace{1pt} Jianbing Shen$^{1}$~\!, \hspace{1pt}  Yu-Wing Tai$^{3}$~\!,\hspace{1pt}  David Crandall$^{4}$~\!, \hspace{1pt} Steven C. H. Hoi$^{5,6}$  \\
	\small{$^1$} \small Inception Institute of Artificial Intelligence, UAE \hspace{2pt} \small{$^2$} \small ETH Zurich, Switzerland \hspace{0pt} \\
	\small{$^3$} \small Tencent \hspace{3pt}	\small{$^4$} \small Indiana University, USA \hspace{3pt} \small{$^5$} \small  Salesforce Research Asia, Singapore \small{$^6$} \small  Singapore Management University, Singapore\\
	{\tt\small \{carrierlxk,wenguanwang.ai\}@gmail.com}\\
	{\tt\small \url{https://github.com/carrierlxk/MuG}}
}

\maketitle

\begin{abstract}
We propose a new method for video object segmentation (VOS) that addresses object pattern learning from
unlabeled videos, unlike most existing methods which rely heavily on extensive annotated data.  We introduce a unified unsupervised/weakly supervised learning framework, called MuG, that comprehensively captures intrinsic properties of VOS at multiple granularities. Our approach can help advance understanding of visual patterns in VOS and significantly reduce annotation burden. With a carefully-designed architecture and strong representation learning ability, our learned
model can be applied to diverse VOS settings, including object-level zero-shot VOS, instance-level zero-shot VOS, and one-shot VOS. Experiments demonstrate promising performance in these settings,
as well as the potential of MuG in leveraging unlabeled data to further improve the segmentation accuracy. 
\end{abstract}
\vspace*{-3pt}
\section{Introduction}
\begin{figure}[t]
	\centering
	\includegraphics[width=0.99\linewidth]{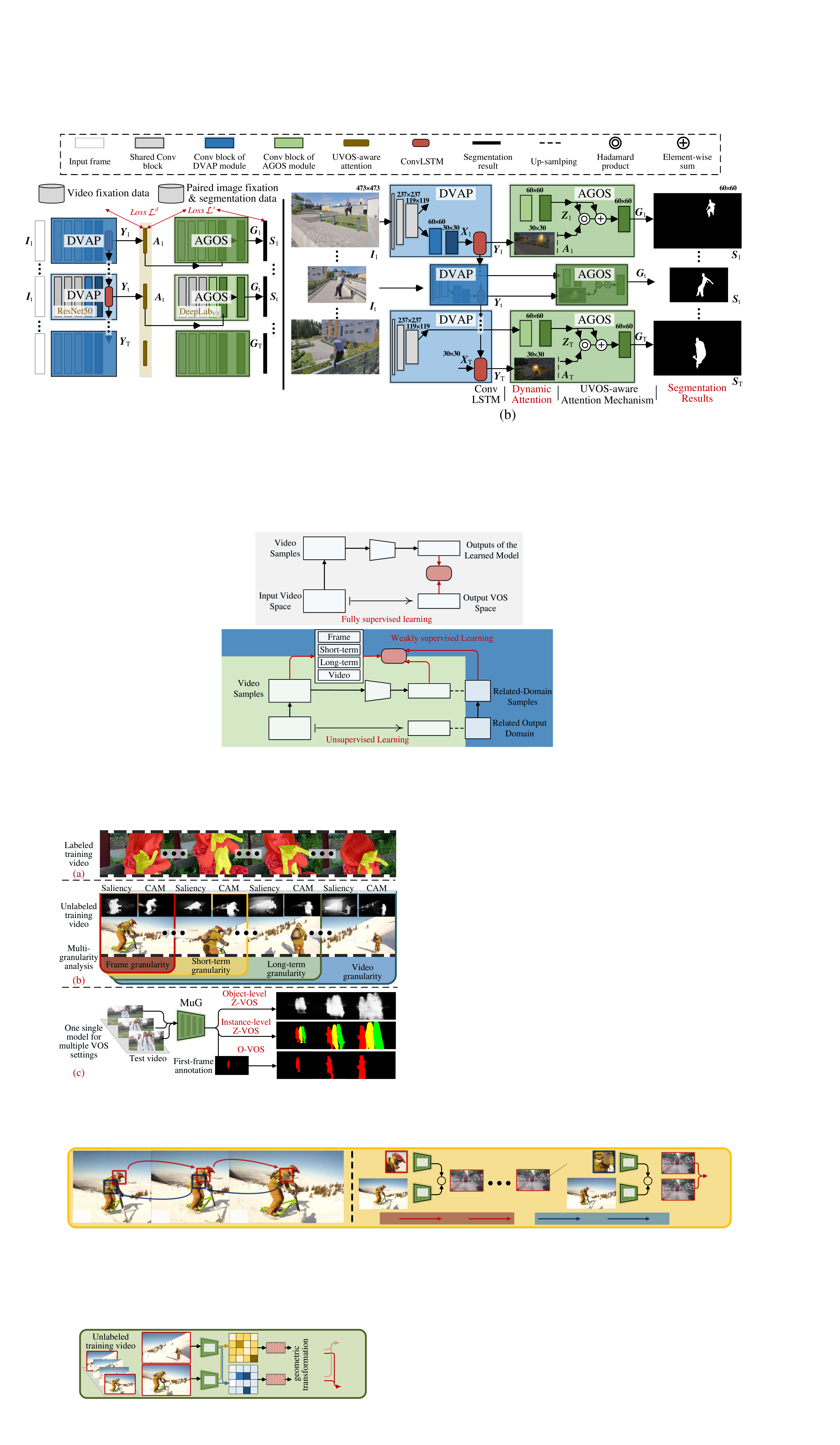}
	\vspace{-9pt}
	\captionsetup{font=small}
	\caption{\small{$_{\!}$(a) Current$_{\!}$ leading$_{\!}$ VOS methods$_{\!}$ are$_{\!}$  learned in a supervised manner,  requiring large-scale elaborately labeled data. (b) Our model, MuG, provides an unsupervised/weakly-supervised framework that learns video object patterns from unlabeled videos. (c) Once trained, MuG can be applied to diverse VOS settings, with strong modeling ability and high generability.
		}
	}
	\label{fig:top-right}
	\vspace{-13pt}
\end{figure}

\vspace{-1pt}
Video object segmentation (VOS) 
has two common settings, zero-shot and one-shot.
Zero-shot VOS (Z-VOS)\footnote{Some conventions~\!\cite{perazzi2016benchmark,DBLP:conf/cvpr/WangSP15} also use `unsupervised VOS' and `semi-supervised VOS' to name the Z-VOS and O-VOS settings~\!\cite{Caelles_2017_CVPR}. In this work, for notational clarity, the terms `supervised', `weakly supervised' and `unsupervised' are only used to address the different learning paradigms. \label{notation}} is to automatically segment out the primary foreground objects, without any test-time human supervision; whereas one-shot VOS (O-VOS) focuses on extracting the human determined foreground objects, typically assuming the first-frame annotations are given ahead inference\textsuperscript{~\!\ref{notation}}. Current leading methods for both Z-VOS and O-VOS are \textit{supervised} deep learning models that require 
extensive amounts of elaborately annotated data to improve the performance and avoid over-fitting. However, obtaining pixel-wise segmentation labels is labor-intensive and expensive (Fig.~\!\ref{fig:top-right}~\!(a)).

It is thus attractive to design VOS models that can learn from unlabeled videos. 
With this aim in mind, we develop a unified, \textit{unsupervised/weakly supervised} VOS method that mines \textit{multi-granularity} cues to facilitate video object pattern learning (Fig.~\!\ref{fig:top-right}~\!(b)). This allows us to take advantage of nearly infinite amounts of video data. 
Below we give a more
formal description of our problem setup and main idea.

\noindent\textbf{Problem Setup and Main Idea.}
Let \bm{$\mathcal{X}$} and \bm{$\mathcal{Y}$} denote the input video space and output VOS space, respectively. Deep learning based VOS solutions seek to learn a differentiable, \textit{ideal} video-to-segment mapping $g^*\!\!\!:\!$ \bm{$\mathcal{X}$}$\mapsto$\bm{$\mathcal{Y}$}. 
To approximate $g^{*\!}$, recent leading VOS models typically work in a \textit{supervised learning} manner, requiring $N$ input
samples and their desired outputs $y_n\!\!:=\!g^{*\!}(x_n)$, where
$\{(x_n,
y_n)\}_{n\!}\!\subset$\bm{$\mathcal{X}$}$\times$\bm{$\mathcal{Y}$}. In contrast, we address the problem in settings with much less supervision: (1) the \textit{unsupervised} case, when we only have samples drawn from  \bm{$\mathcal{X}$}, $\{x_n\}_{n\!}\!\subset$\bm{$\mathcal{X}$}, and want to approximate $g^*$, and (2) the
\textit{weakly supervised learning} setting, in which  we have annotations for
\bm{$\mathcal{K}$}, which  is a related output domain for which obtaining
annotations is easier than \bm{$\mathcal{Y}$}, and we approximate $g^{*}$ using samples from
\bm{$\mathcal{X}$}$\times$\bm{$\mathcal{K}$}. 

The standard way of evaluating learning outcomes follows an \textit{empirical} risk/loss minimization formulation~\!\cite{Shen_2018_CVPR}
:
\vspace{-5pt}
\begin{equation}\small
\tilde{g}\in\mathop{\arg\min}_{g\in\bm{\mathcal{G}}} \frac{1}{N}\sum\nolimits_n\varepsilon(g(x_n),z(x_n)),
\vspace{-0pt}
\end{equation}
where \bm{$\mathcal{G}$} denotes the hypothesis (solution) space, and $\varepsilon\!\!:$\bm{$\mathcal{X}$}$_{\!}$$\times$\bm{$\mathcal{Y}$}$\mapsto\!\!\mathbb{R}$ is an error function that evaluates the estimate $g(x_n)$ against VOS-related prior knowledge $z(x_n)\!\!\in$\bm{$\mathcal{Z}$}. To make $\tilde{g}$  a good approximation of $g^{*\!}$, current supervised VOS methods directly use the desired output $y_n$, \ie, $z(x_n)\!\!:=\!\!g^{*\!}(x_n)$, as the prior knowledge, with the price of vast amounts of well-annotated data.

In our method, the prior knowledge \bm{$\mathcal{Z}$}, in the unsupervised learning setting, is built upon several heuristics and intrinsic properties of VOS itself, while in the weakly supervised learning setting, it additionally considers a related, easily-annotated output domain \bm{$\mathcal{K}$}. For example, part of the fore-background knowledge could be from a saliency model~\!\cite{DBLP:conf/cvpr/YangZLRY13} (Fig.~\!\ref{fig:top-right}~\!(b)), or in a form of CAM maps~\!\cite{zeng2019multi,zhou2016learning} from a pre-trained image classifier~\!\cite{Huang_2017_CVPR} (\ie, a related image classification domain \bm{$\mathcal{K}$})\footnote{\scriptsize{Note that any unsupervised or weakly supervised object segmentation/saliency model can be used; saliency~\!\cite{DBLP:conf/cvpr/YangZLRY13}, and CAM~\!\cite{zeng2019multi,zhou2016learning} are just chosen due to their popularity and relatively high performance.}}. Exploring VOS in an unsupervised or weakly supervised
setting is appealing not only because it  alleviates the annotation burden of
\bm{$\mathcal{Y}$}, but also because it inspires  an in-depth
understanding of the nature of VOS by exploring
\bm{$\mathcal{Z}$}. Specifically, we analyze several different types of cues at multiple granularities, which are crucial for video object pattern modeling:
\vspace*{-0pt}
\begin{itemize}[leftmargin=*]
	\setlength{\itemsep}{0pt}
	\setlength{\parsep}{-2pt}
	\setlength{\parskip}{-0pt}
	\setlength{\leftmargin}{-13pt}
	\vspace{-5pt}
	\item At the \textit{frame} granularity, we leverage information from an unsupervised saliency method~\!\cite{DBLP:conf/cvpr/YangZLRY13} or CAM~\!\cite{zeng2019multi,zhou2016learning} activation maps to enhance the foreground and background discriminability of our intra-frame representation.
	\item At the \textit{short-term} granularity, we impose the local consistency within the representations of short video clips, to describe the continuous and coherent visual patterns within a few seconds.
	\item At the \textit{long-range} granularity, we address semantic correspondence among distant frames, which makes the cross-frame representations robust to local occlusions, appearance variations and shape deformations.
	\item At the \textit{whole-video} granularity, we encourage the video representation to capture global and compact video content, by learning to aggregate multi-frame information and be discriminative to other videos' representations.
	\vspace*{-2pt}
\end{itemize}

All these constraints are formulated under a unified, multi-granularity VOS (MuG) framework, which is fully differentiable and allows unsupervised/weakly supervised video object pattern learning, from unlabeled videos. Our extensive experiments over various VOS settings, \ie, object-level Z-VOS, instance-level Z-VOS, and O-VOS, show that MuG outperforms other unsupervised and weakly supervised methods by a large margin, and continuously improves its performance with more unlabeled data.

\vspace{-3pt}
\section{Related Work}
\vspace{-2pt}

\subsection{Video Object Segmentation}
\vspace{-2pt}
\noindent\textbf{Z-VOS.} As there is no indication for objects to be segmented, conventional ZVOS methods resorted to certain heuristics, such as saliency~\!\cite{DBLP:conf/cvpr/WangSP15,wang2017video,wang2015consistent,DBLP:conf/bmvc/FaktorI14}, object proposals~\!\cite{DBLP:conf/cvpr/KohK17, DBLP:conf/iccv/PerazziWGS15, lee2011key}, and discriminative motion patterns~\!\cite{DBLP:conf/iccv/OchsB11,DBLP:conf/cvpr/FragkiadakiZS12,DBLP:conf/iccv/PapazoglouF13}. 
Recent advances have been driven by deep learning techniques. Various effective networks have been explored, from some early, relatively simple architectures, such as
recurrent network~\!\cite{Song_2018_ECCV,pang2019deep,wang2019learning}, and two-stream network~\!\cite{cheng2017segflow,DBLP:conf/iccv/TokmakovAS17,zhou2020motion}, to recent, more powerful designs, such as teacher-student adaption~\!\cite{siam2018video}, neural co-attention~\!\cite{Lu_2019_CVPR} and graph neural network~\!\cite{wang2019zero,Yan_2019_CVPR}.

\begin{figure*}[b]
	\renewcommand\thefigure{3}
	\vspace{-10pt}
	\centering
	\includegraphics[width=0.99\linewidth]{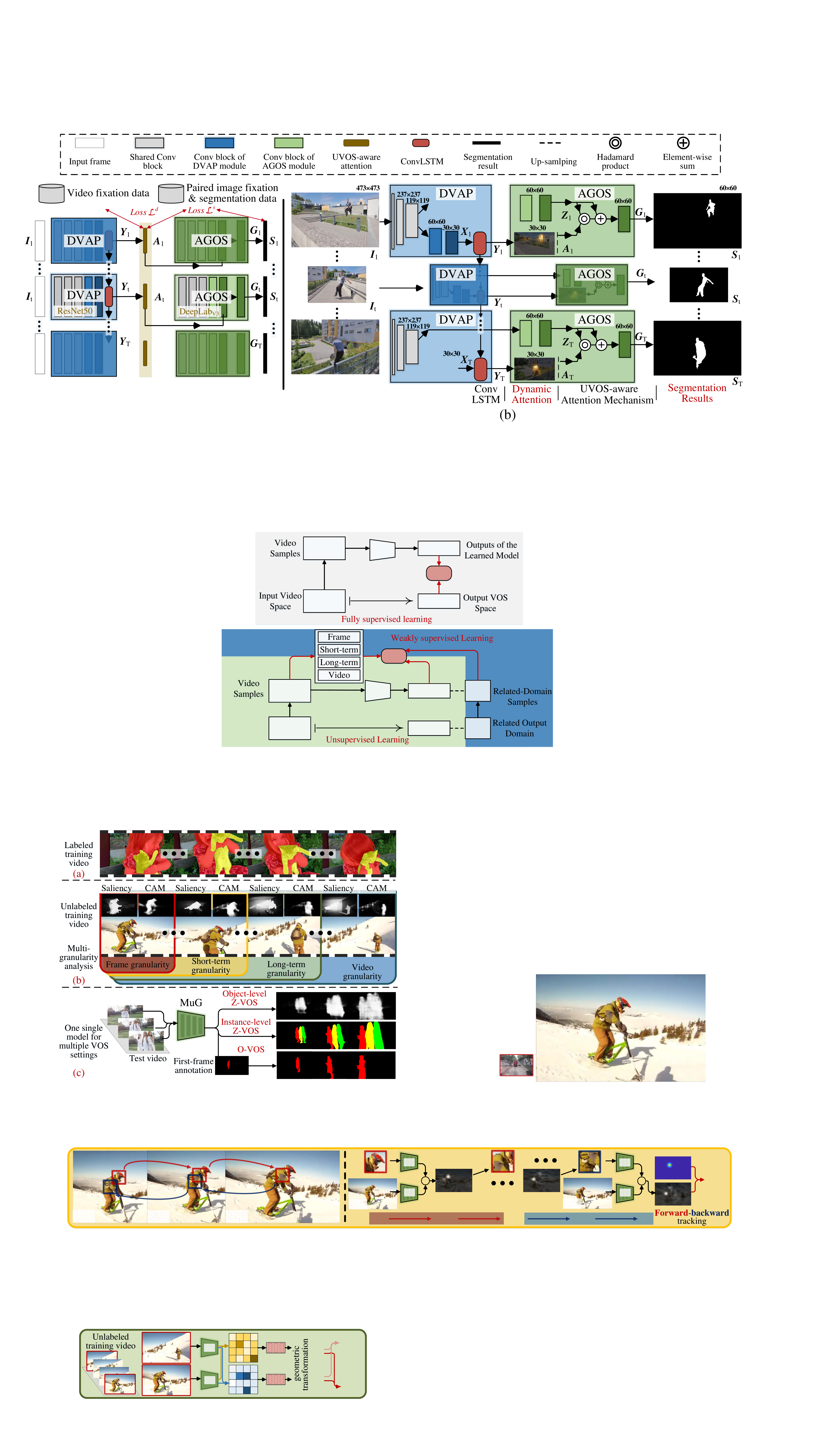}
	\put(-486,7){\scriptsize {$X_{\!t}$}}
	\put(-432,7){\scriptsize {$X_{\!t\!+\!1}$}}
	\put(-374,7){\scriptsize {$X_{\!t\!+\!2}$}}
	\put(-456,47){\small {\textcolor{reda}{$p$}}}
	\put(-396,48){\small {\textcolor{reda}{$p'$}}}
	\put(-333,48){\small {\textcolor{reda}{$p''$}}}
	\put(-283,16.5){\scriptsize {$X_{\!t\!+\!1}$}}
	\put(-279,46.5){\small {\textcolor{reda}{$p$}}}
	\put(-242,26){\scriptsize {$\varphi$}}
	\put(-242,48.5){\scriptsize {$\varphi$}}
	\put(-229.1,33.5){\small {$\star$}}
	\put(-210,21){\scriptsize {$S_{\Rightarrow}$}}
	\put(-265,6){\scriptsize {$X_{\!t}$}}
	\put(-221,6){\scriptsize {$X_{\!t\!+\!1}$}}
	\put(-169,6){\scriptsize {$X_{\!t\!+\!2}$}}
	\put(-119,6){\scriptsize {$X_{\!t\!+\!1}$}}
	\put(-69,6){\scriptsize {$X_{\!t}$}}
	\put(-124,16.5){\scriptsize {$X_{\!t}$}}
	\put(-82,26){\scriptsize {$\varphi$}}
	\put(-82,48.5){\scriptsize {$\varphi$}}
	\put(-69.4,33.5){\small {$\star$}}
	\put(-29,19){\scriptsize {$S_{\Leftarrow}$}}
	\put(-29,48){\scriptsize {$G_p$}}
	\put(-20,34){\scriptsize {\textcolor{reda}{$\mathcal{L}_{\text{short}}$}}}
	\vspace{-8pt}
	\captionsetup{font=small}
	\caption{\small Left: Main idea of short-term granularity analysis. Right: Training details for intra-clip coherence modeling.}
	\label{fig:short-term}
	\vspace{-1pt}
\end{figure*}

\noindent\textbf{O-VOS.} As the annotations for the first frame are assumed available at the test phase, O-VOS focuses on how to accurately propagate the initial labels to subsequent frames. Traditional methods typically used optical flow based propagation strategy~\!\cite{DBLP:conf/cvpr/HicksonBEC14,DBLP:journals/tog/FanZLCC15,wang2018semi,DBLP:conf/cvpr/MarkiPWS16}.  
Now, deep learning based solutions become the main stream, which can be broadly classified into three categories, \ie, \textit{online learning}, \textit{propagation} and \textit{matching} based methods. Online learning based methods~\!\cite{Caelles_2017_CVPR,voigtlaender2017online,DBLP:conf/cvpr/PerazziKBSS17}  fine-tune the segmentation network for each test video on the first-frame annotations. Propagation based methods~\!\cite{DBLP:conf/cvpr/JampaniGG17, wug2018fast,yang2018efficient} rely on the segments of the previous frames and work in a frame-by-frame manner. Matching based methods~\!\cite{wang2019ranet,Voigtlaender_2019_CVPR,luiten2018premvos} segment each frame according to its correspondence/matching relation to the first frame.

Typically, current deep learning based VOS solutions, under either Z-VOS or O-VOS setting, are trained using a large amount of elaborately annotated data for supervised learning.
In contrast, the proposed method trains a VOS network from scratch using unlabeled videos.  This is essential
for understanding how visual recognition works in VOS and for 
narrowing down the annotation budget.


\vspace{-0pt}
\subsection{VOS with Unlabeled Training Videos}
\label{sec:wvos}
\vspace{-1pt}
Learning VOS from unlabeled videos is an essential, yet rarely touched avenue. Among a few efforts,~\cite{pathak2017learning} represents an early attempt in this direction, which uses a modified, purely unsupervised version of~\cite{DBLP:conf/bmvc/FaktorI14} to generate proxy masks as pseudo annotations. 
With a similar spirit, some methods use heuristic segmentation masks~\cite{croitoru2017unsupervised} or weakly supervised location maps~\!\cite{Lee2019FrametoFrameAO} as supervisory signals. With a broader view, some works~\!\cite{Tang_2013_CVPR,hartmann2012weakly,Zhang_2017_CVPR} capitalized on untrimmed videos tagged with semantic labels. In addition to increased annotation efforts, they are hard to handle such a class-agnostic VOS setting. Recently, self-supervised video learning has been applied for O-VOS~\!\cite{vondrick2018tracking,CVPR2019_CycleTime}, which imposes the learned features to capture certain constraints on local coherence, such as cross-frame color consistency~\!\cite{vondrick2018tracking} and temporal cycle-correspondence~\!\cite{CVPR2019_CycleTime}.

Our method is distinctive for two aspects. First, it explores various intrinsic properties of videos as well as class-agnostic fore-background knowledge in a unified, multi-granularity framework, bringing a more comprehensive understanding of visual patterns in VOS. Second, it shows strong video object representation learning ability and, for the first time, it is applied to diverse VOS settings after only being trained once. This gives a new glimpse into the connections between the two most influential VOS settings.

\vspace{-2pt}
\section{Proposed Algorithm}
\vspace{-2pt}
\subsection{Multi-Granularity VOS Network}
\begin{figure}[t]
	\renewcommand\thefigure{2}
	\centering
	\includegraphics[width=0.99\linewidth]{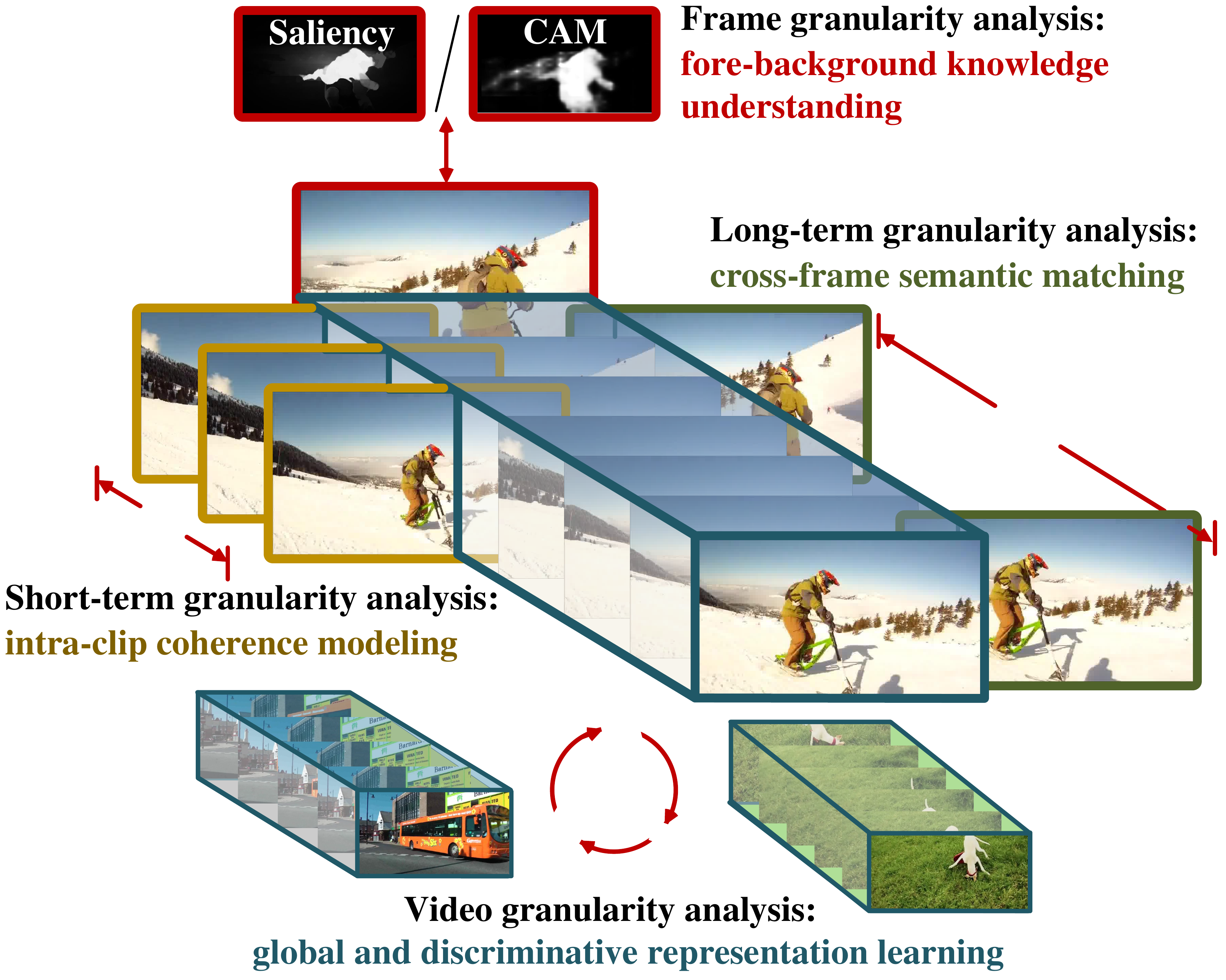}
	\put(-45,106){\footnotesize {\textcolor{reda}{$\mathcal{L}_{\text{long}}$}}}
	\put(-128,36){\footnotesize {\textcolor{reda}{$\mathcal{L}_{\text{global}}$}}}
	\put(-219,86){\footnotesize {\textcolor{reda}{$\mathcal{L}_{\text{short}}$}}}
	\put(-147,157){\footnotesize {\textcolor{reda}{$\mathcal{L}_{\text{frame}}$}}}
	\vspace{-8pt}
	\captionsetup{font=small}
	\caption{\small Overview of our approach. Intrinsic properties over {\color{reda}\textbf{frame}}, {\color{myyellow}\textbf{short-term}}, {\color{mygreen}\textbf{long-term}} and {\color{myblue}\textbf{whole video}} granularities are explored to guide the video object pattern learning.}
	\label{fig:overview}
	\vspace{-12pt}
\end{figure}

\vspace{-1pt}
\label{sec:multi-gran-network}
For a training video $\mathcal{X}\!\!\in$\bm{$\mathcal{X}$} containing $T$ frames: $\mathcal{X}\!=\!\{X_t\}_{t=1}^T$, its features are specified as $\{\textbf{\textit{x}}_{t}\}_{t=1}^T$, obtained from a fully convolutional feature extractor $\varphi$: $\textbf{\textit{x}}_{t\!}\!=_{\!}\!\varphi(X_t)\!\in_{\!}\!\mathbb{R}^{W\!\times\! H\!\times\!C}$. Four-granularity
characteristics are explored to guide the learning of $\varphi$ (Fig.~\!\ref{fig:overview}), described as follows.

\noindent\textbf{Frame Granularity Analysis: Fore-background Knowledge Understanding.} As $\varphi$ is VOS-aware, basic fore-background knowledge is desired to be encoded. In our method, such knowledge (Fig.~\!\ref{fig:top-right}~\!(b)) is initially from a background prior based saliency model~\!\cite{DBLP:conf/cvpr/YangZLRY13} (in an unsupervised learning setting), or in a form of CAM maps~\!\cite{zeng2019multi,zhou2016learning} (in a weakly supervised learning setting).

Formally, for each frame $X_t$, let us denote its corresponding initial fore-background mask as $Q_t\!\in\!\{0,1\}^{W\!\times\!H}$ (\ie, a binarized saliency or CAM activation map). In our frame granularity analysis, the learning of $\varphi$ is guided by the supervision signals of $\{Q_t\}_{t=1}^T$, \ie, utilizing the intra-frame information $\textbf{\textit{x}}_{t\!}\!=_{\!}\!\varphi(X_t)$ to regress $Q_t$:
\vspace{-2pt}
\begin{equation}\small
\begin{aligned}
\mathcal{L}_{\text{frame}}=\mathcal{L}_{\text{CE}}(P_t, Q_t).
\end{aligned}
\vspace{-2pt}
\label{eq:frame}
\end{equation}
Here $\mathcal{L}_{\text{CE\!}}$ is the \textit{cross-entropy} loss, and $P_t\!=\!\rho(\textbf{\textit{x}}_{t})$ where $\rho_{\!}\!:_{\!}\!\mathbb{R}^{\!W\!\times\!H\!\times\!C}\!\!\mapsto\!\![0,1]^{\!W\!\times\!H\!}$ maps the input single-frame feature $\textbf{\textit{x}}_{t}$ into a fore-background prediction map $P_t$. $\rho$ is implemented by a $1\!\times\!1$ convolutional layer with \textit{sigmoid} activation.

\noindent\textbf{Short-Term Granularity Analysis: Intra-Clip Coherence Modeling.} Short-term coherence is an essential property in videos, as temporally close frames typically exhibit continuous visual content changes~\!\cite{hurri2003simple}. To capture this property, we apply a forward-backward patch tracking mechanism~\cite{wang2019unsupervised}. It learns  $\varphi$ by
tracking a sampled patch forwards in a few successive frames and then backwards until the start frame, and penalizing the distance between the initial and final backwards tracked positions of that patch.

Formally, given two consecutive frames $X_{t}$ and $X_{t+1}$, we first crop a patch $p$ from $X_t$ and apply $\varphi$ on $p$ and $X_{t+1}$, separately. Then we get two feature embeddings: $\varphi(p)\!\in\!\mathbb{R}^{w\!\times\!h\!\times\!C\!}$  and $\textbf{\textit{x}}_{t+1\!}\!=_{\!}\!\varphi(X_{t+1})\!\in\!\mathbb{R}^{W\!\times\!H\!\times\!C\!}$. With a design similar to the classic Siamese tracker~\!\cite{bertinetto2016fully}, we forward track the patch $p$ on the next frame $X_{t+1}$ by conducting a cross-correlation operation `$\star$' on $\varphi(p)$ and $\varphi(X_{t+1})$:
\vspace{-2pt}
\begin{equation}\small
S_{\Rightarrow} = \varphi(p)\star\varphi(X_{t+1}) \in [0, 1]^{W\!\times\!H},
\label{equ:fortrack}
\vspace{-1pt}
\end{equation}
where$_{\!}$ $S_{\Rightarrow\!}$ is a \textit{sigmoid}-normalized response map whose size is rescaled into$_{\!}$ $(H_{\!}, W)$. The new location of $p$ in$_{\!}$ $X_{t+1\!}$ is then inferred according to the peak value on$_{\!}$ $S_{\Rightarrow}$. After obtaining the forward tracked patch $p'$ in$_{\!}$ $X_{t+1}$, we backward track$_{\!}$  $p'_{\!}$ to$_{\!}$ $X_{t\!}$ and$_{\!}$  get$_{\!}$  a$_{\!}$  backward tracking response map$_{\!}$ $S_{\Leftarrow}$:
\vspace{-8pt}
\begin{equation}\small
S_{\Leftarrow} = \varphi(p')\star\varphi(X_{t}) \in [0, 1]^{W\!\times\!H}.
\label{equ:backtrack-correlation}
\vspace{-0pt}
\end{equation}
Ideally, the peak of $S_{\Leftarrow\!}$ should correspond to the location of $p$ in the initial frame $X_t$. Thus we build a consistency loss that measures the alignment error
between the initial and forward-backward tracked positions of $p$:
\vspace{-2pt}
\begin{equation}\small
\mathcal{L}_{\text{short}} = \|S_{\Leftarrow}- G_p\|_2^2,
\label{equ:clip-level}
\vspace{-1pt}
\end{equation} 	
where $G_p\!\in\![0, 1]^{W\!\times\!H}$ is a $(H, W)$-dimensional Gaussian-shape map with the same center of $p$ and variance proportional to the size of $p$. As in~\!\cite{wang2019unsupervised}, the above forward-backward tracking mechanism is extended to a multi-frame setting (Fig.~\!\ref{fig:short-term}). Specifically, after obtaining the forward tracked patch $p{'\!}$ in $X_{t+1}$, $p{'\!}$ is further tracked to the next frame$_{\!}$ $X_{t+2}$, and$_{\!}$ a$_{\!}$ new$_{\!}$ tracked$_{\!}$ patch $p''$ is obtained. Then $p''\!$ is reversely tracked$_{\!}$ to$_{\!}$ $X_{t+1\!}$ and$_{\!}$ further$_{\!}$ to$_{\!}$ the$_{\!}$ initial$_{\!}$ frame $X_{t}$, and the local consistency loss in Eq.~\!\ref{equ:clip-level} is computed. Moreover, during training, we first random sample a short video clip consisting of six successive frames. Then we perform above forward-backward tracking based learning strategy over three frames random drawn from the six-frame video clip. With above designs, $\varphi$ captures the spatiotemporally local correspondence and is content-discriminative (due to its cross-frame target re-identification nature).


\begin{figure}[t]
	\centering
	\includegraphics[width=0.99\linewidth]{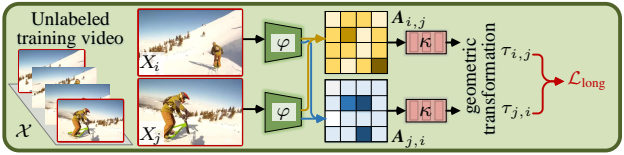}
	\put(-230,6){\scriptsize {$\mathcal{X}$}}
	\put(-183,32){\scriptsize {$X_{\!i}$}}
	\put(-183,5){\scriptsize {$X_{\!j}$}}
	\put(-131,14){\scriptsize {$\varphi$}}
	\put(-131,40){\scriptsize {$\varphi$}}
	\put(-87,49){\scriptsize {$\textbf{\textit{A}}_{i,j}$}}
	\put(-87,4){\scriptsize {$\textbf{\textit{A}}_{j,i}$}}
	\put(-77,40){\footnotesize {$\kappa$}}
	\put(-77,14){\footnotesize {$\kappa$}}
	\put(-46.2,37){\scriptsize {$\tau_{i,j}$}}
	\put(-46.,15){\scriptsize {$\tau_{j,i}$}}
	\put(-21,26){\scriptsize {\textcolor{reda}{$\mathcal{L}_{\text{long}}$}}}
	\vspace{-10pt}
	\captionsetup{font=small}
	\caption{\small Illustration of our long-term granularity analysis.}
	\label{fig:long-term}
	\vspace{-12pt}
\end{figure}

\noindent\textbf{Long-Term Granularity Analysis: Cross-Frame Semantic Matching.} In addition to the local consistency among adjacent frames, there also exist strong semantic correlations among distant frames, as frames from the same video typically contain similar
content~\!\cite{Hossein2009,yan2019fine}. Capturing this property is essential for $\varphi$, as it makes $\varphi$ robust to many challenges, such as appearance variants, shape deformations, object occlusions, \etc. To address this issue, we conduct a long-term granularity analysis, which casts cross-frame correspondence learning as a dual-frame semantic matching problem (Fig.~\!\ref{fig:long-term}). Specifically, given a training pair of two \textit{disordered} frames $(X_i, X_j)$ randomly sampled from $\mathcal{X}$, we compute a similarity affinity $\textbf{\textit{A}}_{i,j}$ between their embeddings: $(\varphi(X_i), \varphi(X_j))$ by a co-attention operation~\!\cite{DBLP:conf/nips/VaswaniSPUJGKP17}:
\vspace{-1pt}
\begin{equation}\small
\textbf{\textit{A}}_{i,j}=\text{softmax}({\textbf{\textit{x}}_i}^{\!\!\!\top}\textbf{\textit{x}}_j)\in [0, 1]^{(W\!H)\times (W\!H)},
\label{equ:long-coattention}
\vspace{-0pt}
\end{equation}
where ${\textbf{\textit{x}}_{i\!}}\!\in_{\!}\!\mathbb{R}^{C\!\times\!(W\!H)\!\!}$ and$_{\!}$ ${\textbf{\textit{x}}_{j\!}}\!\!\in_{\!}\!\mathbb{R}^{C\!\times\!(W\!H)\!}$ are$_{\!}$ flat$_{\!}$ matrix formats of $\varphi(X_i)$ and $\varphi(X_j)$, respectively. `\text{softmax}' indicates \textit{column-wise softmax} normalization.
Given the normalized cross-correlation $\textbf{\textit{A}}_{i,j}$, in line with~\!\cite{rocco2018end}, we use a small neural network $\kappa_{\!}\!:\!\mathbb{R}^{(W\!\times\!H)\!\times\!(W\!\times\!H)\!\!}\!\mapsto\!\mathbb{R}^{6\!}$ to regress the parameters of a geometric transformation $\tau_{i,j}$, \ie, six-degree of freedom (translation, rotation and scale). $\tau_{i,j\!}\!:\!\mathbb{R}^{2}\!\mapsto\!\mathbb{R}^{2}$ gives the relations between the spatial coordinates in $X_i$ and $X_j$ considering the corresponding semantic similarity:
\vspace{-2pt}
\begin{equation}\small
\textbf{\textit{m}}_i=\tau_{i,j}(\textbf{\textit{m}}_j),
\vspace{-1pt}
\end{equation}
where $\textbf{\textit{m}}_i$ is a 2-D spatial coordinate of $X_i$, and $\textbf{\textit{m}}_j$ the corresponding sampling coordinates in $X_j$. Using $\tau_{i,j}$, we can warp $X_i$ to $X_j$.
Similarly, we can also compute $\tau_{j,i}$, \ie, a 2-D warping from $X_j$ to $X_i$. Let us consider two sampling coordinates $\textbf{\textit{m}}_i$ and $\textbf{\textit{n}}_j$ in $X_j$ and $X_i$, respectively, we introduce a semantic matching loss~\!\cite{rocco2018end}:
\vspace{-1pt}
\begin{equation}\small
\begin{aligned}
\mathcal{L}_{\text{long}}=-\big(&\sum\nolimits_{\textbf{\textit{m}}_i\in\Omega}\sum\nolimits_{\textbf{\textit{o}}_j\in\Omega}\textbf{\textit{A}}_{i,j}(\textbf{\textit{m}}_i, \textbf{\textit{o}}_j)\iota(\textbf{\textit{m}}_i, \textbf{\textit{o}}_j)+\\[-2mm] &\sum\nolimits_{\textbf{\textit{n}}_j\in\Omega}\sum\nolimits_{\textbf{\textit{o}}_i\in\Omega}\textbf{\textit{A}}_{j,i}(\textbf{\textit{n}}_j, \textbf{\textit{o}}_i)\iota(\textbf{\textit{m}}_i, \textbf{\textit{o}}_i)\big),
\end{aligned}
\label{equ:semantic-matching}
\vspace{-1pt}
\end{equation}
where $\Omega$ refers to the image lattice, $\textbf{\textit{A}}_{i,j}(\textbf{\textit{m}}_i, \textbf{\textit{o}}_j)\!\in_{\!}\![0,1]$ gives the similarity value between the positions $\textbf{\textit{m}}_i$ and $\textbf{\textit{o}}_j$ in $X_i$ and $X_j$, and $\iota(\textbf{\textit{m}}_i, \textbf{\textit{o}}_j)$ determines if the correspondence between $\textbf{\textit{m}}_i$ and $\textbf{\textit{o}}_j$ is geometrically consistent.$_{\!}$ If$_{\!}$ $||\textbf{\textit{m}}_i, \tau_{i,j}(\textbf{\textit{o}}_j)||_{\!}\!\leq_{\!}\!1$, $\iota_{\!}\!=_{\!}\!1$; otherwise $\iota_{\!}\!=_{\!}\!0$.


\noindent\textbf{Video Granularity Analysis: Global and Discriminative Representation Learning.} 
So far, we have used the pairwise cross-frame information in local and long terms to boost the learning of $\varphi$. $\varphi$ is also desired to learn a compact and globally discriminative video representation. To achieve this, with a \textit{global information aggregation} module, we perform a video granularity analysis within an unsupervised video embedding learning framework~\!\cite{dosovitskiy2014discriminative}, which leverages supervision signals from different videos.

Starting$_{\!}$ with$_{\!}$ our$_{\!}$ global$_{\!}$ information$_{\!}$ aggregation$_{\!}$ module, we split 
$\mathcal{X}_{\!}\!=_{\!}\!\{_{\!}X_{t\!}\}_{t=1\!}^T$ into $K$ segments of equal durations:$_{\!}$ \!$\mathcal{X}_{\!}\!=\!\cup^K_{k=1}\mathcal{X}_k$. For each segment  $\mathcal{X}_{k}$, we randomly sample a single frame, resulting in a $K_{\!}$-frame abstract  $\mathcal{X}'_{\!}\!\!=_{\!}\!\{X_{t_k}\}_{k=1\!}^K$ of $\mathcal{X}$. $\mathcal{X}'_{\!}$ reduces the redundancy among successive frames while preserving global information.

With a similar spirit of \textit{key}-\textit{value} retrieval networks~\!\cite{sukhbaatar2015end}, for each $X_{t_k}\!\in\!\mathcal{X}'$, we set it as a \textit{query} and the rest frames $\mathcal{X}'/X_{t_k}$ as \textit{reference}. Then we compute the normalized cross-correlation between the query and reference:
\vspace{-1pt}
\begin{equation}\small
\!\!\!\!\textbf{\textit{A}}_{t_k\!}\!\!=\!\text{softmax}({\textbf{\textit{x}}_{t_{k}}}^{\!\!\!\!\!\!\top}[\{\textbf{\textit{x}}_{t_{k'}}\}_{t_{k'}}])\in [0,1]^{(W\!H)\times(W\!H\!(K-1))},
\label{equ:global-coattention}
\vspace{-1pt}
\end{equation}
where $k'\!\in\!\{1,\cdots,K\}/k$, and `[$\cdot$]' denotes the concatenation operation. ${\textbf{\textit{x}}_{t_k}}\!\in_{\!}\!\mathbb{R}^{C\!\times\!(W\!H)\!}$
and $[\{\textbf{\textit{x}}_{t_{k'}}\}_{t_{k'}\in\{1,\cdots,K\}/k}]\!\in_{\!}\!\mathbb{R}^{C\!\times\!(W\!H\!(K-1))}$ are flat feature matrices  of the query and reference, respectively. Subsequently, $\textbf{\textit{A}}_{t_k}$ is used as a weight matrix for global information summarization:
\vspace{-1pt}
\begin{equation}\small
\!\!\!\!\textbf{\textit{x}}'_{t_k}\!=\![\{\textbf{\textit{x}}_{t_{k'\!}}\}_{t_{k'\!}}]~\textbf{\textit{A}}_{t_k}^{\!\top}\!\in \! \mathbb{R}^{(W\!H)\!\times\!C\!},~~\text{where~} k'\!\in\!\{1,\cdots\!,K\}/k.\!\!
\label{equ:aggregation}
\vspace{-1pt}
\end{equation}
Our global information aggregation module gathers information from the reference set by a correlation-based feature summarization procedure. For the query frame $X_{t_{k}}$, we obtain its global information augmented representation by:
\vspace{-1pt}
\begin{equation}\small
\textbf{\textit{r}}_{t_k} = [\textbf{\textit{x}}'_{t_k}, \textbf{\textit{x}}_{t_k}] \in \! \mathbb{R}^{W\! \times \!H\!\times \!2C}.
\vspace{-1pt}
\label{eq:global}
\end{equation}

During training, the video granularity analysis essentially discriminates between a set of surrogate video classes~\!\cite{dosovitskiy2014discriminative}. Specifically, given $N$ training videos, we randomly sample a single frame from each video, leading to $N$ training \textit{instances}: $\{X^{n}\}_{n=1}^N$. The core idea is that, for a query frame $X^n_{t_k}$ in the $n$-th video, its global feature embedding is close to the instance $X^{n}$ from the same $n$-th video, and far from other unrelated instances $\{X^{n'}\}_{n'\neq n}$ (from the other$_{\!}$ $N_{\!}\!-_{\!}\!1$ videos). We$_{\!}$ solve$_{\!}$ this$_{\!}$ as$_{\!}$ a$_{\!}$ binary$_{\!}$ classification
problem via maximum likelihood estimation (MLE). In particular, for $X^n_{t_k}$,  instance $X^n$ should be classified into $n$, while other instances $\{X^{n'}\}_{n'\neq n}$ shouldn't be. The probability of
$X^n$ being recognized as instance $n$ is:
\vspace{-3pt}
\begin{equation}\small
P(n|X^n) = \frac{\exp(\text{GAP}(\textbf{\textit{r}}^{n\!\top}_{t_k} \textbf{\textit{r}}^{n}))}{\sum_{i=1}^{N}\exp(\text{GAP}(\textbf{\textit{r}}^{n\!\top}_{t_{k}} \textbf{\textit{r}}^{i}))}.
\vspace{-2pt}
\end{equation}
where `\text{GAP}' stands for \textit{global average pooling}. Similarly, given $X^n_{t_k}$, the probability of other instances $X^{n'}$ be recognized as instance $n$ is:
\vspace{-6pt}
\begin{equation}\small
P(n|X^{n'}) = \frac{\exp(\text{GAP}(\textbf{\textit{r}}^{n\!\top}_{t_k} \textbf{\textit{r}}^{n'}))}{\sum_{i=1}^{N}\exp(\text{GAP}(\textbf{\textit{r}}^{n\!\top}_{t_{k}} \textbf{\textit{r}}^{i}))}.
\vspace{-3pt}
\end{equation}
Correspondingly, the probability of $X^{n'\!}$ not being recognized as instance$_{\!}$ $n$ is$_{\!}$ $1_{\!}\!-_{\!}\!P(n|X^{n'\!})$. The$_{\!}$ joint$_{\!}$ probability of$_{\!}$ $X^{n\!}$ being$_{\!}$ recognized$_{\!}$ as$_{\!}$ instance$_{\!}$ $n_{\!}$ and$_{\!}$ $X^{n'\!}$ not$_{\!}$ being is: $P(n|X^{n})\prod_{n'\neq n}(1\!-\!P(n|X^{n'\!}))$, under the assumption that different instances being recognized as $n$ are independent.

Then the loss function is defined as the negative log likelihood over $N$ query frames from $N$ videos:
\vspace{-1pt}
\begin{equation} \small
\mathcal{L}_{\text{global}}\!=\!-\!\sum_n\log P(n|X^{n})-\!\sum_n\!\sum_{n'\neq n\!}\log(1\!-\!P(n|X^{n'})).
\label{equ:MLE}
\vspace{-3pt}
\end{equation}
Next we will describe the network architecture during the training and inference phases. An appealing advantage of our multi-granularity VOS network is that, after being trained in a unified mode, it can be directly applied to both Z-VOS and O-VOS settings with only slight adaption.

\vspace{-2pt}
\subsection{\!One$_{\!}$ Training$_{\!}$ Phase$_{\!}$ for$_{\!}$ both$_{\!}$ Z-VOS$_{\!}$ and$_{\!}$ O-VOS}
\label{sec:architecture}
\vspace{-1pt}
\noindent\textbf{Network Architecture.} Our whole module is end-to-end trainable.
The video representation space $\varphi$ is learned by a fully convolutional network, whose design is inspired by ResNet-50~\!\cite{he2016deep}. In particular, the first four groups of convolutional layers in ResNet are preserved and dilated convolutional layer~\!\cite{yu2015multi} is used to maintain enough spatial details as well as ensure a large receptive field, resulting in a 512-channel feature representation $\textbf{\textit{x}}$ whose spatial dimensions are $1/4$ of an input video frame $X$.

During training, we use a mini-batch of $N_{\!}\!=_{\!}\!16$ videos and scale all the training frames into $256_{\!}\!\times_{\!}\!256$ pixels.
For frame granularity analysis, 
all the frames access to the supervision signal from the loss $\mathcal{L}_{\text{frame}}$ in Eq.~\!\ref{eq:frame}.

For short-term granularity analysis, six successive video frames are first randomly sampled from each training video, resulting a six-frame video clip. For each video clip, we further sample three video frames orderly and randomly crop a $64\!\times\!64$ patch as $p$. With the feature embedding $\varphi(p)\!\in\!\mathbb{R}^{16\!\times\!16\!\times64}$ of $p$, we forward-backward track $p$ and get its final backward tracking response map ${S}_{\Leftarrow\!}\!\in\![0,1]^{64\!\times\!64}$ via Eq.~\!\ref{equ:backtrack-correlation}. For computing the loss in Eq.~\!\ref{equ:clip-level}, the Gaussian-shape map $G_p\!\in\![0,1]^{64\!\times\!64}$ is obtained by convolving the center position of $p$ with a two-dimension Gaussian map with a kernel width proportional (0.1) to the patch size.

For long-term granularity analysis, after randomly sampling two \textit{disordered} frames $(X_i, X_j)$ ($|i\!-\!j|\!\!\geq\!\!6$) from a training video $\mathcal{X}$, we compute the correlation map $\textbf{\textit{A}}_{i,j}\!\in\![0,1]^{(64\!\times\!64)\!\times\!(64\!\times\!64)}$ by the normalized inner production operation in Eq.~\!\ref{equ:long-coattention}. For the geometric transformation parameter estimator $\kappa_{\!}\!:\!\mathbb{R}^{(64\!\times\!64)\!\times\!(64\!\times\!64)}\!\mapsto\!\mathbb{R}^{6\!}$, it is achieved by two convolutional layers and one linear layer, as in~\!\cite{rocco2018end}. Then the semantic matching loss in Eq.~\!\ref{equ:semantic-matching} is computed.

For video granularity analysis, we split each training video $\mathcal{X}$ into $K\!\!=\!\!8$ segments, and get the global information augmented representation $\textbf{\textit{r}}_{t_k}\!\in\!\mathbb{R}^{64\!\times\!64\!\times\!256}$  for each query frame $X_{t_{k}}$ by Eq.~\!\ref{eq:global}. Then, we compute the soft-max embedding learning loss using Eq.~\!\ref{equ:MLE}, which leverages supervision signals from the $N$ training videos.

\noindent\textbf{Iterative Training by  Bootstrapping.} As seen  in Fig.~\!\ref{fig:top-right}~\!(b), the fore-background knowledge from the saliency~\!\cite{DBLP:conf/cvpr/YangZLRY13} or CAM~\!\cite{zeng2019multi,zhou2016learning} is ambiguous and noisy. Inspired by Bootstrapping~\!\cite{reed2014training}, we apply an iterative training strategy: after training with the initial fore-background maps, we use our trained model to re-label the training data. With each iteration, the learner bootstraps itself by mining better fore-background knowledge and then leading a better model. 
Specifically, for each training frame $X$, given the initial fore-background mask $Q\!\in\!\{0,1\}^{64\!\times\!64}$ and current prediction $\bar{P}^i\!\in\!\{0,1\}^{64\!\times\!64}$ of the model in $i$-th training iteration, the loss in Eq.~\!\ref{eq:frame} in $(i\!+\!1)$-th iteration is formulated in a bootstrapping format:
\vspace{-2pt}
\begin{equation}\small
\begin{aligned}
\!\!\!\!\mathcal{L}^{(i+1)}_{\text{frame}}\!\!=\!\!\sum\nolimits_{\textbf{\textit{m}}\in\Omega}&[\alpha Q_{\textbf{\textit{m}}}\!+\!(1\!-\!\alpha)\bar{P}_{\textbf{\textit{m}}}^{i}]\log(P_{\textbf{\textit{m}}}^{i+1})+\\[-0.7mm] &[\alpha(1\!-\!Q_{\textbf{\textit{m}}})\!+\!(1\!-\!\alpha)(1\!-\!\bar{P}_{\textbf{\textit{m}}}^{i})]\!\log(1\!-\!P_{\textbf{\textit{m}}}^{i+1}),\!\!\!
\end{aligned}
\vspace{-1pt}
\label{eq:framenew}
\end{equation}
where $\alpha\!=\!0.05$ and $Q_{\textbf{\textit{m}}}$ gives the value in position $\textbf{\textit{m}}$. In such a design, the `confident' fore-background knowledge is generated as a convex combination of the initial fore-background information $Q$ and model prediction $P$.

In the $i$-th training iteration, the overall loss to optimize the whole network parameters is the combination of the losses in Eq.~\!\ref{eq:framenew},~\!\ref{equ:backtrack-correlation},~\!\ref{equ:semantic-matching} 
and~\!\ref{equ:MLE}:
\vspace{-2pt}
\begin{equation}\small
\mathcal{L}^{(h)}\!=\!\mathcal{L}^{(h)}_{\text{frame}}\!+\!\beta_1\mathcal{L}_{\text{short}}\!+\! \beta_2\mathcal{L}_{\text{long}}\!+\!\beta_3\mathcal{L}_{\text{global}},
\label{equ:total}
\vspace{-1pt}
\end{equation}
where $\beta$s are coefficients: $\beta_1\!=\!0.1, \beta_2\!=\!0.02$ and $\beta_3\!=\!0.5$. 

The above designs enable a unified un-/weakly supervised feature learning framework. Once the model is trained, the learned representations $\varphi$ can be used for Z-VOS and O-VOS, with slight modifications. In practice, we find that our model can perform well after being trained with 2 iterations; please see~\!\S\ref{sec:ablation} for related experiments.


\vspace{-2pt}
\subsection{Inference Modes for Z-VOS and O-VOS}
\label{sec:inference}
\vspace{-1pt}
Now we detail our inference modes for object-level Z-VOS, instance-level Z-VOS, and O-VOS settings.

\noindent\textbf{Object-Level Z-VOS Setting.} For each test frame, object-level Z-VOS aims to predict a binary segmentation mask where the primary foreground objects are separated from the background while the identities of different foreground objects are not distinguished. In the classic VOS setting, since there is no any test-time human intervention, how to discover the primary video objects is the central problem. Considering the fact that interested objects frequently appear throughout the video sequence, we readout the segmentation results from the global information augmented feature $\textbf{\textit{r}}$, instead of directly using intra-frame information to predict the fore-background mask (\ie, $\rho(\textbf{\textit{x}})$).
This is achieved by an extra segmentation readout layer $\upsilon_{\!}\!:\!\mathbb{R}^{64\!\times\!64\!\times\!256}\!\mapsto\![0,1]^{64\!\times\!64}$, which takes the global frame embedding $\textbf{\textit{r}}$ as the input and produces the final object-level segmentation prediction. $\upsilon$ is also trained by the \textit{cross-entropy} loss, as in Eq.~\!\ref{eq:framenew}. For notation clarity, we omit this term in the overall training loss in Eq.~\!\ref{equ:total}.  Please note that $\upsilon$ is only used in Z-VOS setting; for O-VOS setting, the segmentation masks are generated with a different strategy.

\noindent\textbf{Instance-Level Z-VOS Setting.} Our model can also be
adapted for the instance-level Z-VOS setting, in which different
object instances must be discriminated, in addition to separating the
primary video objects from the background without test-time human
supervision.  For each test frame, we first apply mask-RCNN~\!\cite{he2017mask} to produce a set of category agnostic object proposals.
Then we apply our trained model for producing a binary foreground-background mask per frame.
After combining object bounding-box proposals with binary object-level segmentation masks,
we can filter out the background proposals and obtain pixel-wise, instance-level object candidates for each frame. Finally, to link those object candidates across different frames, similar to~\!\cite{luiten2018premvos}, we use overlap ratio and optical flow as the cross-frame candidate-association metric. Note that, mask-RCNN can be replaced with non-learning Edgebox~\cite{ZitnickECCV14edgeBoxes} and GrabCut, resulting a purely unsupervised/weakly-supervised protocol.

\noindent\textbf{O-VOS Setting.} In O-VOS, for each test video sequence, instance-level annotations regarding multiple general foreground objects in the first frame are given.  In such a setting, our trained network works in a per-frame matching based mask propagation fashion. Concretely, assume there are a total of $L$ object instances (including the background) in the first-frame annotation, each spatial position $\textbf{\textit{n}}_{\!}\!\in_{\!}\!\Omega$ will be associated with a one-hot class vector $\hat{\textbf{\textit{y}}}_{\textbf{\textit{n}}\!}\!\in_{\!}\!\{0,1\}^L$, whose element $\hat{y}^l_{\textbf{\textit{n}}\!}$ indicates whether pixel $\textbf{\textit{n}}_{\!}$ belong to $l$-th object instance. Starting from the second frame, we use both the last segmented frame $X_{t-1}$ as well as current under-segmented frame $X_{t}$ to build an input pair for our model. Then we compute their similarity affinity $\textbf{\textit{A}}_{t-1,t\!}\!\in_{\!}\![0, 1]^{(64\!\times\!64)\!\times\!(64\!\times\!64)}$ in the feature space: $\textbf{\textit{A}}_{t-1,t\!}\!=_{\!}\!\text{softmax}({\textbf{\textit{x}}_{t-1}}^{\!\!\!\top}\textbf{\textit{x}}_{t})$. After that, for each pixel $\textbf{\textit{m}}$ in $X_{t}$, we compute its probability distribution $\textbf{\textit{v}}_{\textbf{\textit{m}}}\!\in\![0,1]^L$ over the $L$ object instances as:
\vspace{-2pt}
\begin{equation}\small
\textbf{\textit{v}}_{\textbf{\textit{m}}}\!=\!\sum\nolimits_{\textbf{\textit{n}}\in\Omega}\textbf{\textit{A}}_{t-1,t}(\textbf{\textit{n}},\textbf{\textit{m}})~~\hat{\textbf{\textit{y}}}_{\textbf{\textit{m}}},
\vspace{-1pt}
\end{equation}
where $\textbf{\textit{A}}_{t-1,t}(\textbf{\textit{n}}, \textbf{\textit{m}})\!\in_{\!}\![0,1]_{\!}$ is the affinity value between pixel $\textbf{\textit{n}}$ in $X_{t-1}$ and $\textbf{\textit{m}}$ in $X_t$.  For $\textbf{\textit{m}}$, it is assigned to $l^{*\!}$-th instance: $l^{*\!}\!=_{\!}\!\mathop{\arg\max}_l(\{v^l_{\textbf{\textit{m}}}\}_{l=1}^L)$, where $\textbf{\textit{v}}_{\textbf{\textit{m}}\!}\!=\![v^l_{\textbf{\textit{m}}}]_{l=1}^L$. Then we get its label vector $\hat{\textbf{\textit{y}}}_{\textbf{\textit{m}}}$. In this way, from the segmented frame $X_{t}$, we move to the next input frame pair $(X_{t},X_{t+1})$ and get the segmentation result for $X_{t+1}$. As our method does not use any first-frame fine-tuning~\!\cite{cheng2017segflow,DBLP:conf/cvpr/PerazziKBSS17} or online learning~\cite{voigtlaender2017online} technique, it is fast for inference.

\begin{table}[t]
	\centering
	\resizebox{0.49\textwidth}{!}{
		\setlength\tabcolsep{1pt}
		\renewcommand\arraystretch{1.0}
		\begin{tabular}{c|c||cc||cc}
			\hline\thickhline
			\rowcolor{mygray}
			&	 &\multicolumn{2}{c||}{Unsuper.} &\multicolumn{2}{c}{Weakly-super.}\\
			\rowcolor{mygray}
			\multirow{-2}{*}{Aspects} &\multirow{-2}{*}{Module}  &mean $\mathcal{J}$ & $\Delta$$\mathcal{J}$ &mean $\mathcal{J}$ & $\Delta$$\mathcal{J}$\\
			\hline
			\hline 
			Reference&\textbf{Full model} (2 iterations)  & 58.0 &- & 61.2 &-\\
			\hline
			\hline
			\multirow{2}{*}{\tabincell{c}{Initial Fore-/Background\\Knowledge}}
			&Heuristic Saliency~\!\cite{DBLP:conf/cvpr/YangZLRY13} & 37.2&-20.8 &- &-\\
			&CAM~\!\cite{zeng2019multi} &- &- &45.3 &-15.9\\
			\hline
			\multirow{4}{*}{\tabincell{c}{Multi-Granularity\\Analysis}}
			&\textit{w/o.}  Frame Granularity  & 40.2 & -17.8 & 40.2  & -21.0\\
			&\textit{w/o.}  Short-term Granularity  & 51.3 & -6.7 & 57.1 & -4.1\\
			&\textit{w/o.}  Long-term Granularity  & 52.8 & -5.2 & 56.0 &-5.2 \\
			&\textit{w/o.}  Video  Granularity  & 56.4 & -1.6 & 60.4 & -0.8\\
			\hline
			\multirow{3}{*}{\tabincell{c}{Iterative Training\\ via Bootstrapping}} & 1 iteration~ & 50.8 & -7.2 & 54.9 & -6.3\\
			& 3 iterations & 58.0 & 0.0 & 61.2 & 0.0\\
			& 4 iterations & 58.0 & 0.0 & 61.2 & 0.0\\
			\hline
			More Data& 	+ LaSOT dataset~\!\cite{Fan_2019_CVPR}  & 59.5 & +1.5& 62.3 & +1.1\\
			\hline	
			Post-Process& 	\textit{w/o.} CRF  & 55.3 & -2.7 & 58.7 & -2.5\\
			\hline
	\end{tabular}}	
	\vspace*{-8pt}
	\captionsetup{font=small}
	\caption{\small \textbf{Ablation study on DAVIS$_{16}$~\!\cite{perazzi2016benchmark} \texttt{val} set}, under the object-level Z-VOS setting. Please see~\!\S\ref{sec:ablation} for details.}
	\label{table:abl}
	\vspace*{-12pt}	
\end{table}

\vspace{-3pt}
\section{Experiment}
\label{sec:exp}
\vspace{-2pt}
\subsection{Common Setup}
\vspace{-1pt}
\noindent\textbf{Implementation Details.}  We train the whole network from scratch on the OxUvA~\!\cite{valmadre2018long} tracking dataset, as in~\!\cite{Lai19}. OxUvA comprises 366 video sequences with more than 1.5 million frames in total. We train our model with  SGD optimizer. For our bootstrapping based iterative training, two iterations are used and each takes about 8 hours.


\begin{table*}[t]
	\centering
	\begin{threeparttable}
		\resizebox{0.99\textwidth}{!}{
			\setlength\tabcolsep{6pt}
			\renewcommand\arraystretch{1.0}
			\begin{tabular}{lc||cccccc||ccc||cccccc}
				\hline\thickhline
				\rowcolor{mygray}
				\multicolumn{2}{c||}{Supervision}&\multicolumn{6}{c||}{Non Learning} & \multicolumn{3}{c||}{Unsupervised Learning} & \multicolumn{2}{c}{Weakly-supervised}\\
				\rowcolor{mygray}
				\multicolumn{2}{c||}{Method}&TRC~\!\cite{DBLP:conf/cvpr/FragkiadakiZS12}&CVOS~\!\cite{DBLP:conf/cvpr/TaylorKS15}&KEY~\!\cite{lee2011key} &MSG~\!\cite{DBLP:conf/iccv/OchsB11} &NLC~\!\cite{DBLP:conf/bmvc/FaktorI14} &FST~\!\cite{DBLP:conf/iccv/PapazoglouF13}  &Motion Masks~\!\cite{pathak2017learning} &TSN~\!\cite{croitoru2017unsupervised}  &\textbf{Ours}&COSEG~\!\cite{tsai2016semantic} &\textbf{Ours}\\
				\hline
				\hline
				\multirow{3}{*}{ $\mathcal{J}$ } & Mean~ $\uparrow$ & 47.3&48.2   &  49.8& 53.3& 55.1 & \textbf{55.8} &{48.9}&31.2&\textbf{58.0}& 52.8 & \textbf{61.2}\\
				& Recall $\uparrow$  & 49.3&54.0& 59.1& 61.6& 55.8 &\textbf{64.7} &44.7 &18.7 &\textbf{65.3} &50.0 & \textbf{65.9}  \\
				& Decay~$\downarrow$    & 8.3&10.5&14.1& 2.4 & 12.6 &\textbf{0.0} &19.2& \textbf{-0.4}&{2.0} &\textbf{10.7} &11.6 \\\hline
				\multirow{3}{*}{ $\mathcal{F}$ } & Mean~ $\uparrow$ &44.1 & 44.7 &  42.7 & 50.8 & \textbf{52.3}  & 51.1   &39.1 & 18.4 &  \textbf{51.5}&49.3 &   \textbf{56.1} \\
				& Recall $\uparrow$  &43.6&52.6& 37.5&60.0 &  \textbf{51.9}&51.6&28.6 &5.6 &\textbf{53.2} &52.7 & \textbf{54.6} \\	
				& Decay~$\downarrow$     &12.9& 11.7&10.6 & 5.1& 11.4 &\textbf{2.9}&17.9&\textbf{1.9}	&2.1& \textbf{10.5} & 20.3\\ \hline
				$\mathcal{T}$  & Mean~ $\downarrow$   & {39.1}& \textbf{25.0}& 26.9 &30.1  &42.5& 36.6 &36.4& 37.5& \textbf{30.1} & 28.2&\textbf{23.6}\\
				\hline
			\end{tabular}
		}
	\end{threeparttable}
	\vspace*{-8pt}
	\captionsetup{font=small}
	\caption{\small \textbf{Evaluation of object-level Z-VOS on DAVIS$_{16}$ \texttt{val} set~\!\cite{perazzi2016benchmark} (~\!\S\ref{sec:OZVOS})}, with region similarity $\mathcal{J}$, boundary accuracy $\mathcal{F}$ and time stability $\mathcal{T}$. 
		(The best scores in each supervision setting are marked in \textbf{bold}. These notes are the same to other tables.) }
	\label{OZVOSdavis16}
	\vspace*{-8pt}	
\end{table*}

\begin{table*}[t]
	\centering
	\begin{threeparttable}
		\resizebox{0.99\textwidth}{!}{
			\setlength\tabcolsep{5pt}
			\renewcommand\arraystretch{1.0}
			\begin{tabular}{lc||cccc||ccc||cccccc}
				\hline\thickhline
				\rowcolor{mygray}
				\multicolumn{2}{c||}{Supervision}&\multicolumn{4}{c||}{Non Learning} & \multicolumn{3}{c||}{Unsupervised Learning} & \multicolumn{4}{c}{Weakly-supervised Learning}\\
				\rowcolor{mygray}
				\multicolumn{2}{c||}{Method}&CRANE~\!\cite{Tang_2013_CVPR} &NLC~\!\cite{DBLP:conf/bmvc/FaktorI14}&FST~\!\cite{DBLP:conf/iccv/PapazoglouF13} &ARP~\!\cite{DBLP:conf/cvpr/KohK17}   &Motion Masks~\!\cite{pathak2017learning} &TSN~\!\cite{croitoru2017unsupervised}  &\textbf{Ours}&SOSD~\!\cite{zhang2015semantic}&BBF~\!\cite{Fatemeh2017semantic}&COSEG~\!\cite{tsai2016semantic} &\textbf{Ours}\\
				\hline
				\hline
				{ $\mathcal{J}$ } Mean~$\uparrow$& & 23.9&27.7 &  \textbf{53.8}& 46.2& 32.1& 52.2 & \textbf{57.7} &54.1&53.3&58.1& \textbf{62.4}\\
				
				\hline
			\end{tabular}
		}
	\end{threeparttable}
	\vspace*{-8pt}
	\captionsetup{font=small}
	\caption{\small \textbf{Evaluation of object-level Z-VOS on Youtube-Objects~\!\cite{prest2012learning} (\S\ref{sec:OZVOS})}, with mean $\mathcal{J}$. See the supplementary for more details.	}
	\label{OZVOSYoutube}
	\vspace*{-16pt}	
\end{table*}

\noindent\textbf{Configuration and Reproducibility.}  MuG is implemented on PyTorch. All experiments are conducted on an Nvidia TITAN Xp GPU and an Intel (R) Xeon E5 CPU. All our implementations, trained models, and segmentation results will be released to provide the full details of our approach.

\vspace{-3pt}
\subsection{Diagnostic Experiments}
\label{sec:ablation}
\vspace{-1pt}
A series of ablation studies are performed for assessing the effectiveness of each essential component of MuG.

\noindent\textbf{Initial Fore-Background Knowledge.}
Baselines \textit{Heuristic Saliency} and \textit{CAM} give the scores of initial fore-background knowledge, based on their CRF-binarized outputs. As seen, with the low-quality initial knowledge, our MuG gains huge performance improvements ($+20.8\%$ and $+15.9\%$ promotions), showing the significance of our multi-granularity video object pattern learning scheme.

\noindent\textbf{Multi-Granularity Analysis.}
Next we  investigate the contributions of multi-granularity cues in depth. As shown in Table~\!\ref{table:abl}, the intrinsic, multi-granularity properties are indeed meaningful, as disabling any granularity analysis component causes performance to erode. For instance, removing the frame granularity analysis during learning hurts performance (mean $\mathcal{J}$: $\!58.0\!\rightarrow\!40.2$, $\!61.2\!\rightarrow\!40.2$), due to the lack of fore-/background information. Similarly, performance drops when excluding short- or long-term granularity analysis, suggesting the importance of capturing local consistency and semantic correspondence. Moreover, considering video granularity information also improves the final performance, proving the meaning of comprehensive video content understanding in video object pattern modeling.

\noindent\textbf{Iterative Training Strategy.}
From Table~\!\ref{table:abl}, we can see that with more iterations of our bootstrapping training strategy
($1\!\rightarrow\!2$), better performance can be obtained. However, further iterations ($2\!\rightarrow\!4$) give only marginal performance change. We thus use two iterations in all the experiments.

\noindent\textbf{More Training Data.} To show the potential of our unsupervised/weakly supervised VOS learning scheme, we probe the upper bound by training on additional videos. With more training data (1400 videos) from LaSOT dataset~\!\cite{Fan_2019_CVPR}, performance boosts can be observed in both two settings.

\vspace{-2pt}
\subsection{Performance for Object-Level Z-VOS}
\label{sec:OZVOS}
\vspace{-1pt}
\noindent\textbf{Datasets.} Experiments are conducted on two famous Z-VOS datasets: DAVIS~\!\cite{perazzi2016benchmark} and Youtube-Objects~\!\cite{prest2012learning}, which have pixel-wise, object-level annotations. DAVIS$_{16}$ has 50 videos (3,455 frames), covering a wide range of challenges, such as fast motion, occlusion, dynamic background, \etc. It is split into a \texttt{train} set (30 videos) and a \texttt{val} set (20 videos).
Youtube-Objects contains 126 video sequences that belong to 10 categories (such as \textit{cat}, \textit{dog}, \etc) and has 25,673 frames in total. The \texttt{val} set of DAVIS$_{16}$ and whole Youtube-Objects are used for evaluation.

\noindent\textbf{Evaluation Criteria.}  For fair comparison, we follow the official
evaluation protocols of each dataset. For DAVIS$_{16}$, we report region similarity $\mathcal{J}$, boundary accuracy $\mathcal{F}$ and time stability $\mathcal{T}$.  For Youtube-Objects,  the performance is evaluated in terms of region similarity $\mathcal{J}$.

\noindent\textbf{Post-processing.} Following the common protocol in this area~\!\cite{DBLP:conf/iccv/TokmakovAS17,Song_2018_ECCV,cheng2017segflow},  the final segmentation results are optimized by CRF~\!\cite{krahenbuhl2011efficient}  (about 0.3s per frame).

\noindent\textbf{Quantitative Results.} 	Table~\!\ref{OZVOSdavis16} presents the comparison results with several  non-learning, unsupervised or weakly supervised learning competitors in DAVIS$_{16}$ dataset. In particular, MuG exceeds current leading unsupervised learning-based methods (\ie, Motion Masks~\!\cite{pathak2017learning} and TSN~\!\cite{croitoru2017unsupervised} ) in large margins (58.0 \textit{vs} 48.9 and 58.0 \textit{vs} 31.2). MuG also outperforms classical weakly-supervised Z-VOS method COSEG~\!\cite{tsai2016semantic}, and all the previous heuristic methods.  Table~\!\ref{OZVOSYoutube} summarizes comparison results on Youtube-Objects dataset, showing again our superior performance in both unsupervised and weakly supervised learning settings. 

\noindent\textbf{Runtime Comparison.} The inference time of MuG is about 0.6s per frame, which is faster than most deep learning based competitors (\eg, MotionMask~\!\cite{pathak2017learning} (1.1s), TSN~\!\cite{croitoru2017unsupervised} (0.9s)). This is because, except CRF~\!\cite{krahenbuhl2011efficient}, there is no other pre-/post-processing step (\eg, superpixel~\!\cite{tsai2016semantic}, optical flow~\!\cite{DBLP:conf/iccv/PapazoglouF13}, \etc) and online fine-tuning~\!\cite{DBLP:conf/cvpr/KohK17}.

\begin{table}[t]
	\centering
	\vspace*{5pt}	
	\begin{threeparttable}
		\resizebox{0.49\textwidth}{!}{
			\setlength\tabcolsep{4pt}
			\renewcommand\arraystretch{1.0}
			\begin{tabular}{cc||ccc||cc||cc}
				\hline\thickhline
				\rowcolor{mygray}
				\multicolumn{2}{c||}{Supervision}&\multicolumn{3}{c||}{Fully Supervised} & \multicolumn{2}{c||}{Unsupervised} & \multicolumn{2}{c}{Weakly-super.}\\
				\rowcolor{mygray}
				\multicolumn{2}{c||}{}&AGS&{PDB}& RVOS &&&&\\
			\rowcolor{mygray}
				\multicolumn{2}{c||}{\multirow{-2}{*}{Method}}& ~\!\cite{wang2019learning} &~\!\cite{Song_2018_ECCV} & ~\!\cite{ventura2019rvos} &  \multirow{-2}{*}{\textbf{Ours*}}& \multirow{-2}{*}{\textbf{Ours}} &\multirow{-2}{*}{\textbf{Ours*}}& \multirow{-2}{*}{\textbf{Ours}} \\
				\hline
				\hline
				$\mathcal{J}\&\mathcal{F}$ & Mean~ $\uparrow$  & 45.6& 40.4&22.5&36.5 &37.3& 40.6&41.7 \\
				\hline
				\multirow{3}{*}{ $\mathcal{J}$ } & Mean~ $\uparrow$    &  42.1 & 37.7& 17.7& 33.8&35.0& 37.7&38.9  \\
				& Recall $\uparrow$   & 48.5& 42.6& 16.2&38.2 &39.3&42.5&44.3  \\
				
				& Decay~$\downarrow$    & 2.6& 4.0 & 1.6&2.1 &3.8&1.9 &2.7  \\\hline
				\multirow{3}{*}{ $\mathcal{F}$ } & Mean~ $\uparrow$     &49.0 & 43.0 & 27.3&38.0&39.6&43.5&44.5 \\			
				& Recall $\uparrow$  & 51.5&44.6& 24.8 & 38.6 &41.1 &44.9&46.6\\	
				& Decay~$\downarrow$     &2.6 & 3.7 &1.8& 3.2&4.6&1.0&1.7  \\
				\hline
			\end{tabular}
		}
	\end{threeparttable}
	\captionsetup{font=small}
	\vspace*{-8pt}
	\caption{\small \textbf{Evaluation of instance-level Z-VOS on DAVIS$_{17}$ \texttt{test-dev} set~\!\cite{Caelles_arXiv_2019} (\S\ref{sec:IZVOS})}, $*$ denotes purely unsupervised/weakly-supervised protocol with non-learning Edgebox~\cite{ZitnickECCV14edgeBoxes} and GrabCut.
	}
	\label{IZVOSdavis17}
	\vspace*{-13pt}	
\end{table}

\begin{table*}[t]
	\centering
	\begin{threeparttable}
		\resizebox{0.99\textwidth}{!}{
			\setlength\tabcolsep{2pt}
			\renewcommand\arraystretch{1.0}
			\begin{tabular}{lc||ccccc||ccccc||ccc}
				\hline\thickhline
				\rowcolor{mygray}
				\multicolumn{2}{c||}{Supervision}&\multicolumn{5}{c||}{Non Learning} & \multicolumn{5}{c||}{Unsupervised Learning} & \multicolumn{2}{c}{Weakly-supervised}\\
				\rowcolor{mygray}
				\multicolumn{2}{c||}{Method}&HVS~\!\cite{DBLP:conf/cvpr/HicksonBEC14}&JMP~\!\cite{DBLP:journals/tog/FanZLCC15} &FCP~\!\cite{DBLP:conf/iccv/PerazziWGS15} &SIFT Flow~\!\cite{liu2010sift}&BVS~\!\cite{DBLP:conf/cvpr/MarkiPWS16}  &Vondrick \etal~\!\cite{vondrick2018tracking}  &mgPFF~\!\cite{kong2019multigrid} &TimeCycle~\!\cite{CVPR2019_CycleTime} &CorrFlow~\!\cite{Lai19}  &\textbf{Ours}&FlowNet2~\!\cite{DBLP:conf/cvpr/IlgMSKDB17}&\textbf{Ours}\\
				\hline
				\hline
				\multirow{3}{*}{ $\mathcal{J}$ } & Mean~ $\uparrow$ &54.6  &  57.0& 58.4&51.1 & \textbf{60.0} & 38.9&40.5 & 55.8 &{48.9}&\textbf{63.1}& 41.6&\textbf{65.7}\\
				& Recall $\uparrow$  & 61.4& 62.6& \textbf{71.5}&58.6 &66.9 & 37.1& 34.9&64.9&44.7 &\textbf{71.9}& 45.7 &\textbf{77.6}  \\
				& Decay~$\downarrow$    &23.6& 39.4&\textbf{ -2.0} &18.8& 28.9& 22.4& 18.8&\textbf{0.0}&19.2& 28.1 & \textbf{19.9}& 26.4 \\\hline
				\multirow{3}{*}{ $\mathcal{F}$ } & Mean~ $\uparrow$ & 52.9 &  53.1 & 49.2 & 44.0&\textbf{58.8} & 30.8& 34.0& 51.1  & 39.1 &\textbf{61.8}& 40.1 &\textbf{63.5}  \\
				& Recall $\uparrow$&61.0& 54.2&49.5 &50.3&\textbf{67.9}&21.7 & 24.2&51.6 &28.6 &\textbf{64.2}& 38.3& \textbf{67.7} \\	
				& Decay~$\downarrow$     & 22.7 & 38.4& \textbf{-1.1} &20.0& 21.3 &16.7 &13.8 &\textbf{2.9}&17.9&30.5 &\textbf{26.6} &27.2	\\ \hline
				$\mathcal{T}$  & Mean~ $\downarrow$   & 36.0& \textbf{15.9} &30.6&16.4&34.7&45.9&53.1&36.6&\textbf{36.4}&43.0& \textbf{29.8} &44.4 \\
				\hline
			\end{tabular}
		}
	\end{threeparttable}
	\captionsetup{font=small}
	\vspace*{-8pt}
	\caption{\small \textbf{Evaluation of O-VOS on DAVIS$_{16}$ \texttt{val} set~\!\cite{perazzi2016benchmark} (\S\ref{sec:OVOS})}, with region similarity $\mathcal{J}$, boundary accuracy $\mathcal{F}$ and time stability $\mathcal{T}$. }
	\label{OVOSdavis16}
	\vspace*{-8pt}	
\end{table*}

\begin{table*}[t]\small
	\centering
	\begin{threeparttable}
		\resizebox{0.99\textwidth}{!}{
			\setlength\tabcolsep{5pt}
			\renewcommand\arraystretch{1.0}
			\begin{tabular}{cc||cc||ccccccc||cccc}
				\hline\thickhline
				\rowcolor{mygray}
				\rowcolor{mygray}
				\multicolumn{2}{c||}{Supervision}&\multicolumn{2}{c||}{Non Learning} & \multicolumn{7}{c||}{Unsupervised Learning} & \multicolumn{2}{c}{Weakly-supervised}\\
				\rowcolor{mygray}
				\multicolumn{2}{c||}{} &SIFT Flow&BVS&DeepCluster&Transitive Inv &Vondrick \etal &mgPFF &TimeCycle &CorrFlow & &FlowNet2& \\
				\rowcolor{mygray}
				\multicolumn{2}{c||}{\multirow{-2}{*}{Method}}&~\!\cite{liu2010sift}&~\!\cite{DBLP:conf/cvpr/MarkiPWS16}& ~\!\cite{caron2018deep}&~\!\cite{wang2017transitive}& ~\!\cite{vondrick2018tracking} &~\!\cite{kong2019multigrid} &~\!\cite{CVPR2019_CycleTime}&~\!\cite{Lai19}&\multirow{-2}{*}{\textbf{Ours}}&~\!\cite{DBLP:conf/cvpr/IlgMSKDB17}&\multirow{-2}{*}{\textbf{Ours}} \\			
				\hline
				\hline
				$\mathcal{J}\&\mathcal{F}$ & Mean~ $\uparrow$  &34.0& \textbf{37.3} &35.4 &29.4&34.0& 44.6 &42.8 &50.3 &\textbf{54.3}&26.0&\textbf{56.1}\\
				\hline
				\multirow{2}{*}{ $\mathcal{J}$ } & Mean~ $\uparrow$ & \textbf{33.0}&32.9&37.5 & 32.0& 34.6& 42.2& 43.0 & 48.4&\textbf{52.6}&26.7&\textbf{54.0}  \\
				& Recall $\uparrow$   & -& 31.8 &- &-&34.1 &41.8 &43.7&53.2&\textbf{57.4}&  23.9&\textbf{60.7} \\
				\hline
				\multirow{3}{*}{ $\mathcal{F}$ } & Mean~ $\uparrow$ &35.0&\textbf{41.7}&33.2 & 26.8&32.7&46.9 &
				42.6& 52.2&\textbf{56.1}&25.2& \textbf{58.2} \\
				& Recall $\uparrow$  &-&41.4&- &-&26.8 & 44.4& 41.3
				&56.0&\textbf{58.1}&24.6& \textbf{62.2}
				\\
				\hline
			\end{tabular}
		}
	\end{threeparttable}
	\captionsetup{font=small}
	\vspace*{-8pt}
	\caption{\small \textbf{Evaluation of O-VOS on DAVIS$_{17}$ \texttt{val} set~\!\cite{pont20172017} (\S\ref{sec:OVOS})}, with region similarity $\mathcal{J}$, boundary accuracy $\mathcal{F}$ and average of $\mathcal{J}\&\mathcal{F}$.
	}
	\label{OVOSdavis17}
	\vspace*{-15pt}	
\end{table*}

\vspace{-2pt}
\subsection{Performance for Instance-Level Z-VOS}
\label{sec:IZVOS}
\vspace{-1pt}
\noindent\textbf{Datasets.} We test the performance for instance-level Z-VOS on DAVIS$_{17}$~\!\cite{Caelles_arXiv_2019} dataset, which has 120 videos and 8,502 frames in total. It has three subsets, namely, \texttt{train},
\texttt{val}, and \texttt{test-dev}, containing 60, 30, and 30 video sequences, respectively. We use the ground-truth masks provided by the newest DAVIS challenge~\!\cite{Caelles_arXiv_2019}, as the original annotations are biased towards the O-VOS scenario.

\noindent\textbf{Evaluation Criteria.}
Three standard evaluation metrics, provided by DAVIS$_{17}$, are used, \ie, region similarity $\mathcal{J}$, boundary accuracy $\mathcal{F}$ and the average value of $\mathcal{T}\&\mathcal{F}$.

\noindent\textbf{Quantitative Results.} Three top-performing ZVOS methods from the DAVIS$_{17}$ benchmark are included. As shown in Table~\!\ref{IZVOSdavis17},  our model achieves comparable performance with the fully supervised methods (\ie, AGS~\!\cite{wang2019learning} and PDB~\!\cite{Song_2018_ECCV}). Notably, it significantly outperforms recent RVOS~\!\cite{ventura2019rvos}  (mean $\mathcal{T}\&\mathcal{F}$: $+14.8\%$ and $+19.2\%$ in unsupervised and weakly-supervised learning setting, respectively).


\noindent\textbf{Runtime Comparison.} The processing time for each frame is about 0.7s which is comparable to AGS~\!\cite{wang2019learning} and PDB~\!\cite{Song_2018_ECCV}, and slightly slower than RVOS~\cite{ventura2019rvos} (0.3s).

\vspace{-3pt}
\subsection{Performance for O-VOS}
\label{sec:OVOS}
\vspace{-3pt}
\noindent\textbf{Datasets.} DAVIS$_{16}$~\!\cite{perazzi2016benchmark} and DAVIS$_{17}$~\!\cite{pont20172017} datasets are used for performance evaluation under the O-VOS setting.

\noindent\textbf{Evaluation Criteria.}
Three standard evaluation criteria are reported: region similarity $\mathcal{J}$, boundary accuracy $\mathcal{F}$ and the average value of $\mathcal{T}\&\mathcal{F}$. For DAVIS$_{16}$ dataset, we further report the time stability $\mathcal{T}$.

\noindent\textbf{Quantitative Results.} Table~\!\ref{OVOSdavis16} and Table~\!\ref{OVOSdavis17} give evaluation results on DAVIS$_{16}$ and DAVIS$_{17}$
, respectively. Table~\!\ref{OVOSdavis16} shows that our unsupervised method exceeds  representative self-supervised methods (\ie, TimeCyle~\!\cite{CVPR2019_CycleTime} and CorrFlow~\!\cite{CVPR2019_CycleTime}) and the best non-learning method (\ie, BVS~\!\cite{DBLP:conf/cvpr/MarkiPWS16}) across most metrics. In particular, with the learned CAM as supervision, our weakly supervised method further improves the performance, \eg, mean $\mathcal{J}$ of 65.7. Table~\ref{OVOSdavis17} verifies again our method performs favorably against the current best unsupervised method, CorrFlow, according to mean $\mathcal{T}\&\mathcal{F}$ (54.3 \textit{vs} 50.3). Note  that CorrFlow and our method use the same training data. This demonstrates our MuG is able to learn more powerful video object patterns, compared to previous self-learning counterparts.

\begin{figure}[t]
	\centering
	\includegraphics[width=.48\textwidth]{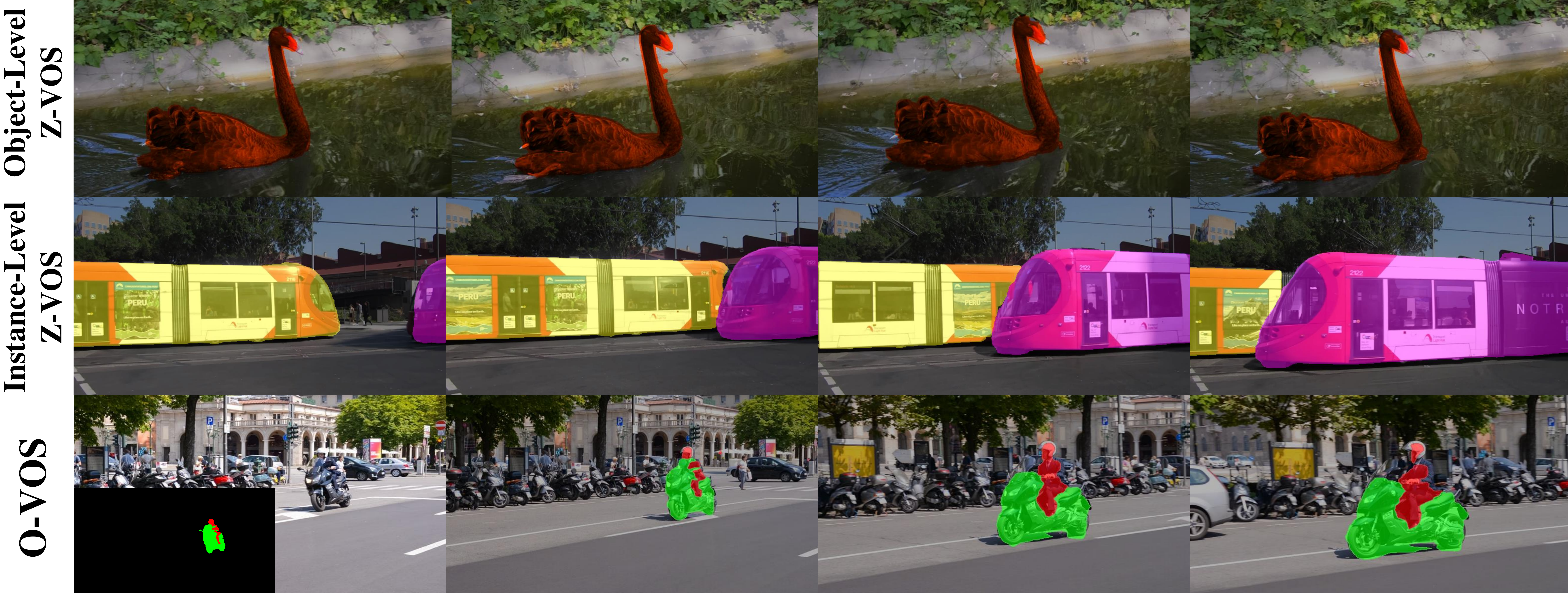}
	\vspace*{-18pt}	
	\captionsetup{font=small}
	\caption{\small Visual results on three videos (top: \textit{blackswan}, middle: \textit{tram}, bottom: \textit{scooter-black}) under object-level Z-VOS, instance-level Z-VOS and O-VOS setting,  respectively (see \!\S\ref{sec:visualization}). For \textit{scooter-black}, its first-frame annotation is also depicted.}
	\label{fig:visualization}	
	\vspace*{-18pt}	
\end{figure}

\noindent\textbf{Runtime Comparison.}
In instance-level Z-VOS setting, MuG runs about 0.4s per frame. This is faster than matching based methods (\eg, SIFT Flow~\!\cite{liu2010sift} (5.1s) and mgPFF~\!\cite{kong2019multigrid} (1.3s)), and favorably against self-supervised learning methods, \eg, TimeCycle~\!\cite{CVPR2019_CycleTime} and CorrFlow~\!\cite{Lai19}. 
\vspace{-3pt}
\subsection{Qualitative Results}
\label{sec:visualization}
\vspace{-2pt}
Fig.~\!\ref{fig:visualization} presents some visual results for object-level ZVOS (top row), instance-level Z-VOS (middle row) and O-VOS (bottom row).
For \textit{blackswan} in DAVIS$_{16}$~\!\cite{perazzi2016benchmark}, the primary objects undergo  view changes and background clutter, but our MuG still generates accurate foreground segments. The effectiveness of instance-level Z-VOS can be observed in \textit{tram} of DAVIS$_{17}$~\!\cite{Caelles_arXiv_2019}. In addition, MuG can produce high-quality results with the given first-frame annotations in O-VOS setting (see the results on the last row for \textit{scooter-black} in DAVIS$_{17}$~\!\cite{pont20172017}), although the different instances suffer from fast motion and scale variation. More results can be found in supplementary materials.

\vspace{-4pt}
\section{Conclusion}
\vspace{-4pt}
We proposed MuG -- an end-to-end trainable, unsupervised/weakly supervised learning approach for segmenting objects from the videos. Different from current popular supervised VOS solutions  requiring extensive amounts of elaborately annotated training samples, our MuG models video object patterns by comprehensively exploring supervision signals from different granularities of unlabeled videos. Our model sets new state-of-the-arts over diverse VOS settings, including object-level Z-VOS, instance-level Z-VOS, and O-VOS. Our model opens up the probability of learning VOS from nearly infinite amount of unlabeled videos and unifying different VOS settings from a single view of video object pattern understanding. 

{\small
	\bibliographystyle{ieee_fullname}
	\bibliography{egbib}
}

\end{document}